\documentclass[sigconf,screen]{acmart}
\setcopyright{none}
\renewcommand\footnotetextcopyrightpermission[1]{}
\usepackage{graphicx}
\usepackage{amsmath,amsthm,amsfonts}
\usepackage{bm}
\usepackage{bbm}
\usepackage{url}
\usepackage{array}
\usepackage{color,soul}
\usepackage{multirow}
\usepackage{booktabs}
\usepackage{enumitem}
\usepackage{subcaption}
\usepackage[ruled,vlined,linesnumbered]{algorithm2e}
\usepackage{setspace}
\usepackage[english]{babel}
\usepackage{makecell}
\usepackage{courier}
\usepackage{bytefield}
\usepackage[utf8]{inputenc}
\usepackage{tikz}
\usepackage{tcolorbox}
\usetikzlibrary{calc}

\theoremstyle{plain}

\newtheorem{rem}{Remark}
\newtheorem{sty1}{Theorem}
\newtheorem{defi}[sty1]{Definition}

\allowdisplaybreaks[4]

\tcbuselibrary{skins, breakable}
\newtcolorbox{promptbox}[1][]{
    colback=gray!5!white,
    colframe=gray!75!black,
    title=\textbf{Prompt Template},
    fonttitle=\bfseries,
    breakable,
    boxrule=0.5mm,
    #1
}

\begin{document}
\title{AI Tool Discovery at Scale: All You Need is DNS}

\author{Enhao Chen}
\affiliation{%
  \institution{The University of Hong Kong}
  \city{Hong Kong}
  \country{China}}
\email{aruleuer@gmail.com}

\author{Yulin Shao}
\affiliation{%
  \institution{The University of Hong Kong}
  \city{Hong Kong}
  \country{China}}
\email{ylshao@hku.hk}

\thanks{The ToolDNS framework and the dataset constructed in this work are open-sourced and publicly available at \url{https://github.com/hku-icl/ToolDNS.git}.}

\begin{abstract}
The coming era of autonomous AI agents demands a discovery mechanism capable of navigating millions of tools, yet existing solutions buckle under $\mathcal{O}(N)$ complexity and centralized governance. Instead of building another fragile overlay, we propose ToolDNS, a radical framework that retrofits semantic tool discovery onto the Internet's most resilient substrate: the Domain Name System (DNS). By embedding functional intent and organizational trust into a hierarchical namespace, ToolDNS transforms an expensive semantic search into a series of lightweight, $\mathcal{O}(\log N)$ name resolutions. We introduce three protocol-compliant enhancements to enable decentralized governance and semantic pruning: partially unfolded names, EDNS0 intent payloads, and logical subdomains.
To rigorously evaluate this approach across the fragmented tooling landscape, we construct and release a large-scale heterogeneous benchmark comprising $33,688$ real-world tools spanning MCP, A2A, RESTful, and Skill protocols. On this dataset, ToolDNS slashes the per-query search space by $95.26\%$ while matching state-of-the-art retrieval accuracy. Furthermore, its UDP-native design reduces discovery latency by orders of magnitude compared to HTTP-based registries. Our work demonstrates that scalable AI interoperability requires not more middleware, but a smarter utilization of the infrastructure already beneath our feet.
\end{abstract}

\maketitle
\thispagestyle{plain}
\pagestyle{plain}


\section{Introduction}

The rapid proliferation of AI agents marks the dawn of a new computational paradigm: isolated models are evolving into collaborative societies where autonomous agents invoke each other's tools such as functions, APIs, skills, or entire agentic workflows \cite{swe,react,shao2024theory}. At the heart of this evolution lies a fundamental yet under-appreciated problem: service discovery \cite{toolformer,gorilla,cui2024llmind}. For an agent to find and invoke the most suitable tool among millions or billions of candidates, the ecosystem requires a discovery mechanism that is scalable, secure, incrementally deployable, and agnostic to protocols (e.g., MCP \cite{mcp_survery}, A2A \cite{a2a_survey}, RESTful API \cite{RestfulAPI}, Skill \cite{skill_servery}).

Existing solutions fall into two broad categories, both of which introduce ad hoc layers on top of the OSI application layer (Layer 7), what one might call Layer 8 or 9 constructs.
\begin{itemize}[leftmargin=0.5cm]
    \item The first category, represented by ToolLLM \cite{toolllm} and similar vector-retrieval systems \cite{a2a_survey}, builds centralized registries or global indexes. For every query, the system computes similarity against all known tools, incurring $\mathcal{O}(N)$ computational cost and unacceptable latency at scale. 
    \item The second category, embodied by frameworks like OpenClaw \cite{openclaw_docs,openclaw_survery}, injects all tool descriptions into the large language model \cite{LLM_survery} (LLM)'s context window. This strategy becomes infeasible when the tool set grows beyond a few hundred entries, as the context window is exhausted and inference cost grows quadratically.
\end{itemize}

\begin{figure}[t]
    \centering
    \includegraphics[width=0.45\textwidth]{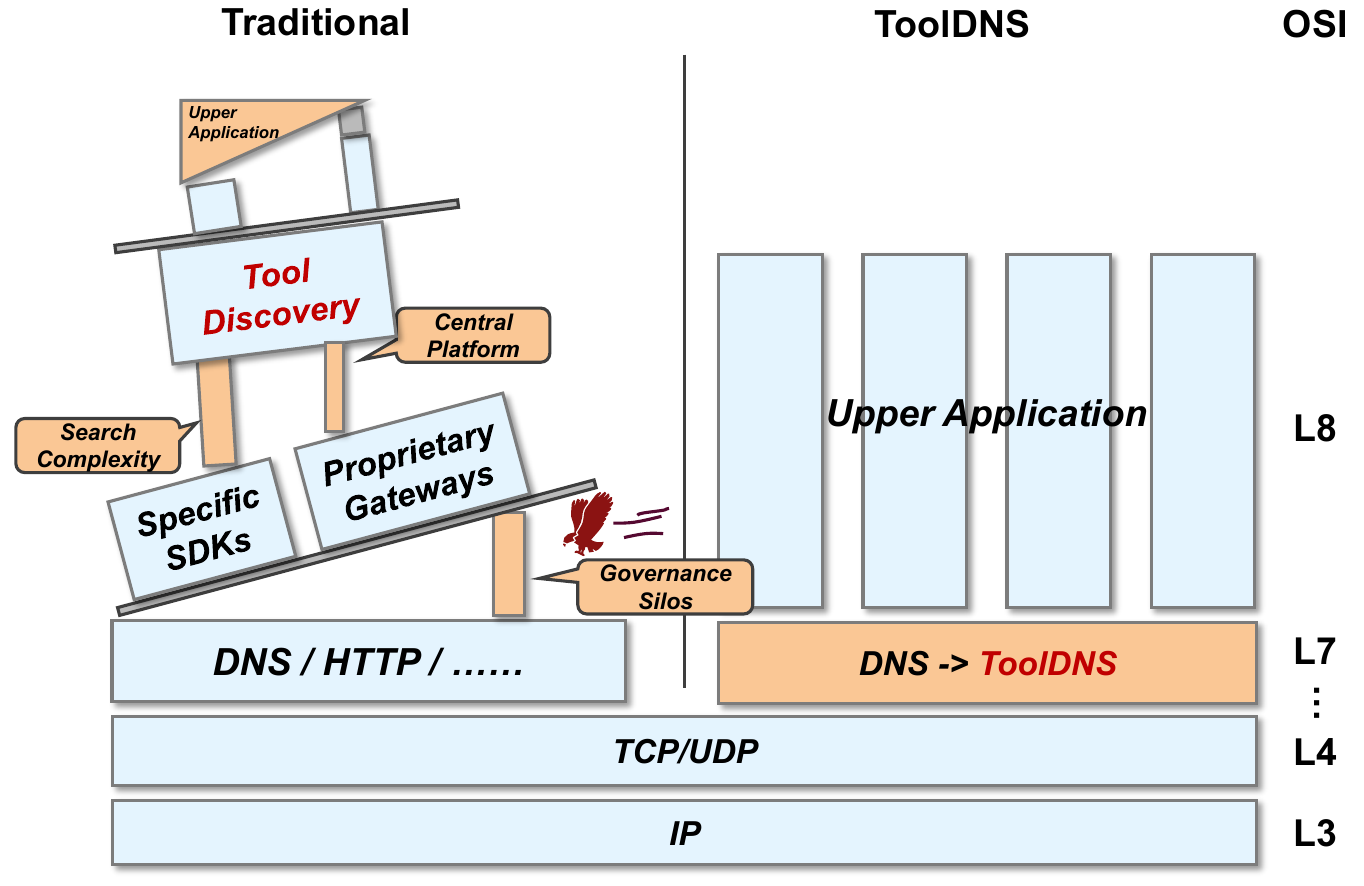}
    \caption{Existing discovery paradigms (left) versus the DNS‑native approach of ToolDNS (right).}
    \Description{Existing discovery paradigms (left) versus the DNS‑native approach of ToolDNS (right).}
    \label{fig:related_work}
\end{figure}

Beyond performance, both approaches suffer from single points of failure, governance silos (different organizations cannot agree on a single registry), and high deployment barriers: any new discovery service requires building and operating a new global infrastructure from scratch, and often mandates integration of a specific SDK or proprietary API for any client seeking to access it. 
In essence, the community risks falling into an over-engineering cycle, creating ever-heavier layer structures while ignoring the robust, globally deployed substrate that already exists \cite{endtoend,lean_software,why_internet_works,shao2021federated,stoica2021cloud}. Fig.~\ref{fig:related_work} contrasts this trend with our proposed direction.

This paper begins with a simple but radical re-examination: \emph{Service discovery, when viewed through the lens of infrastructure design, can be largely reduced to a naming problem rather than purely a semantic matching problem}. Specifically, by embedding functional and trust attributes into a hierarchical namespace, the act of finding a suitable tool becomes equivalent to resolving a domain name. 
When an agent asks ``which tool can translate English to Chinese?'', it is effectively asking for the name of a service that satisfies a set of constraints. 
The Domain Name System (DNS), the most successful and scalable naming system ever built, naturally provides exactly such a capability: its hierarchical namespace, recursive resolution, distributed caching, cryptographic extensions (e.g., {DNSSEC \cite{dnssec}}), and delegation \cite{rfc2181} (NS records) exhibit a profound semantic isomorphism with the requirements of large-scale tool discovery, including hierarchical classification, efficient retrieval, load distribution, verifiable trust, and decentralized governance. 
Yet, prevailing research has dismissed DNS as ``only for static hostnames'', overlooking the latent extensibility encoded in standards like EDNS0 \cite{rfc_edns0} and SRV \cite{RFC_SRV} records.
This oversight has led the community to repeatedly build registration systems that are brittle and lack interoperability.

We propose ToolDNS, a framework that unlocks AI tool discovery by reusing, rather than replacing, the existing DNS infrastructure. ToolDNS operates as a hierarchical system that mirrors the organizational reality: a designated authority maintains an offical domain such as \path{.tools} and \path{weather.tools}, while any other institution (e.g., a university, a company, a standards body) can apply for and manage its own independent subdomain under it, such as \path{hku.weather.tools} or \path{google.weather.tools}.
Tool registration is delegated to these authoritative zones via standard DNS NS records, and agents discover tools through the existing recursive resolver tree. There are no new servers, no central authority, and no mandatory SDKs. Furthermore, based on the connectionless design of DNS, ToolDNS relies solely on UDP packets, which can reduce network overhead and transmission pressure.

However, repurposing DNS for semantic discovery is not straightforward. 
First, standard DNS was designed for exact, deterministic lookups, and clients must know the exact domain name beforehand. AI agents, by contrast, express only vague intent (e.g., find a weather API). 
Second, according to RFC 2181 \cite{rfc2181}, each part of the domain name can be up to $63$ characters long, and the entire domain name cannot exceed $253$ characters, far too short for natural language queries.
Third, in existing DNS, a single entity controls each subdomain, creating governance monopolies that conflict with open multi-organization tool ecosystems. 

ToolDNS overcomes these mismatches not by extending DNS with heavy middleware, but by three lightweight, protocol-compliant enhancements: 
1) partially unfolded domain names, which turn the unknown target into a progressively narrowing search cursor; 
2) EDNS0, which supports significant large length of to carry semantic intents payloads without breaking label limits; and 3) logical subdomains, which decouple the functional hierarchy from administrative control, allowing \path{hku.weather.tools} and \path{google.weather.tools} to coexist under the same \path{weather.tools} branch, each with independent governance. These innovations transform DNS from a static mapper into a semantic discovery framework, without modifying a single line of resolver code.

With these enhancements, ToolDNS inherits all the native advantages of DNS: no new infrastructure to deploy, zero configuration on the client side (requiring only EDNS0 support), native caching of directory structures, and incremental deployability. 
Moreover, it addresses the trust and governance gap directly.
By allowing agents to query only specific organizational subdomains (e.g., only \path{*.hku.tools}), ToolDNS enables flexible security policies without a central trust anchor. 
The framework is also protocol agnostic: MCP, A2A, RESTful API, Skills all fit into the same \path{_service._proto} encoding, providing a lightweight, extensible metadata scheme.

In summary, this work makes the following contributions:
\begin{itemize}[leftmargin=0.5cm]
    \item We put forth ToolDNS, an infrastructure-native hierarchical architecture for AI tool discovery. By encoding functional, domain, and protocol attributes into a semantic namespace under the \path{.tools} TLD, we retrofit tool discovery onto the existing DNS hierarchy. This design bypasses the traditional path of rebuilding centralized registries at a new layer 8 or 9, achieving low engineering migration costs, bidirectional compatibility with standard DNS clients and resolvers, and global high availability inherited from the DNS infrastructure itself. The hierarchical structure reduces per-query search complexity from $O(N)$ to $O(\log N)$.
    \item We introduce a delegated trust and governance mechanism based on logical subdomains and standard DNS NS records. By inheriting the hierarchical endorsement of parent domains, subdomains provide verifiable identity and security governance through the established reputations of authoritative institutions. At the same time, this mechanism circumvents the administrative monopoly of traditional DNS by allowing multiple entities to manage their own independent resources under shared functional namespaces. Agents can restrict discovery queries to trusted subdomains only, achieving decentralized trust management, flexible security isolation, and equal-footing multi-tenant governance without a central authority.
    \item We design an LLM-augmented, index-free retrieval protocol using EDNS0 extensions. Instead of maintaining a global vector index, our method performs on-the-fly semantic pruning: the recursive resolver carries the user's natural language intent and a $K$ parameter to each authoritative server, which returns the top-$K$ most relevant subdomains via lightweight LLM scoring. This avoids periodic index retraining and naturally accommodates dynamic tool repositories. 
    New tools simply appear as new entries in the lowest‑level subdomains and are discovered at query time, eliminating synchronization lag. This protocol turns each DNS iteration into an adaptive, semantic narrowing step, making the discovery process independent of the underlying tool repository's update frequency.
    \item We create and release a large-scale heterogeneous benchmark dataset comprising $33,688$ real-world tools spanning MCP, A2A, RESTful, and Skill protocols. Using this dataset, we empirically validate that ToolDNS reduces the per-query search space by $95.26\%$ through only two layers of hierarchical pruning, while achieving retrieval accuracy comparable to state-of-the-art vector retrieval baselines. Furthermore, ToolDNS improves response latency by orders of magnitude compared to exhaustive scan methods, demonstrating its ability to balance high precision and low overhead in ultra-large-scale deployment scenarios.
\end{itemize}

\section{Problem Formulation}
This section formalizes the AI tool discovery problem, establishes the metrics that capture its inherent challenges, and characterizes the properties an ideal solution must possess.

\subsection{AI Tool Discovery}
Let $\mathcal{T} = \{t_1, t_2, \dots, t_N \}$ denote a set of $N$ tools. Each tool $t$ represents an invocable capability exposed by an AI agent or service. Every tool is associated with:
\begin{itemize}[leftmargin=0.5cm]
\item A functional description $d(t)$, typically a natural language string that specifies what the tool does, its input/output schema, and usage constraints.
\item A protocol specification $p(t) \in \mathcal{P}$, where $\mathcal{P}$ is the set of supported invocation protocols (e.g., MCP, A2A, RESTful, Skill).
\item An access endpoint $e(t)$, typically a network address (IP and port) or a URI.
\item An organizational trust anchor $a(t)$, denoting the entity that publishes and vouches for the tool (e.g., HKU, Google).
\end{itemize}

An agent query $q$ is a natural language utterance expressing the agent's intent to discover and invoke a tool. For example, $q =$ ``fetch the historical weather data for Hong Kong for the past week''.

\begin{defi}
A discovery system is a function that, given an agent query $q$, returns a tool $\tilde t(q)\in \mathcal{T}$
considered relevant. In practice, the system may return a ranked list of size $K$ (top-$K$ discovery), denoted $\mathcal{D}_K(q)\subseteq \mathcal{T}$.
\end{defi}

To evaluate the precision of the discovery system, we further define a relevance set $R_t$ for each tool $t \in \mathcal{T}$. This set $R_t \subseteq \mathcal{T}$ comprises all tools that to the same semantic category as $t$. In a hierarchical namespace, $R_t$ typically corresponds to all tool records sharing the same leaf subdomain.

To measure the discovery quality, we define two metrics.

\begin{defi}[Hit rate]
Given a query $q$ and a ground-truth tool $t^*$ that correctly satisfies $q$ (e.g., the tool the agent should invoke), the hit rate is $\tilde t(q) \subseteq R_{t^*}$
. When averaged over a query set, it measures the fraction of queries for which the correct tool appears in the top-$K$ results.
\end{defi}

\begin{defi}[Queried tool count]
For a query $q$, the queried tool count $C(q)$ is the number of distinct tools whose metadata or representations are actively examined (e.g., scored, compared, or retrieved) during the discovery process. A well-designed discovery system should be achieve to keep $C(q)$ sublinear in $N$:
\begin{equation}
    \lim_{N\to\infty} \frac{C(q)}{N} = 0.
\end{equation}
In particular, $C(q)=\mathcal{O}(\log N)$ or $\mathcal{O}(1)$ is desirable.
\end{defi}

Beyond these quantitative performance metrics, a practical discovery system for open AI agent ecosystems must also satisfy the following properties.
\begin{itemize}[leftmargin=0.5cm]
    \item The system must operate over the existing Internet infrastructure without requiring new global services, custom SDKs, or modifications to client network stacks beyond widely available features \cite{endtoend}.
    \item No single entity should have unilateral control over tool registration or discovery \cite{centrial_invisible}. Different organizations must be able to manage their own tool namespaces independently, while still interoperating under a common logical root.
    \item Agents must be able to enforce security policies (e.g., ``only use tools endorsed by my organization or by well-known security auditors'') without relying on a single global trust anchor.
    \item The discovery mechanism must not depend on the specific invocation protocol of the tool \cite{protocol_decople}. Tools using MCP, A2A, REST, or future protocols must be discoverable through the same interface.
    \item Tools appear, disappear, and change their descriptions frequently. The discovery system should reflect these changes with low latency (ideally, at query time) and without expensive global re-indexing.
\end{itemize}

\subsection{Limitations of Existing Paradigms}
With the metrics and desired properties defined above, we can examine the limitations of current AI tool discovery approaches. These approaches can be grouped into several families, each of which struggles to simultaneously satisfy the requirements of scalability, deployability, decentralized governance, and dynamic adaptability.

\textbf{Vector-based retrieval.}
Schemes such as ToolLLM \cite{toolllm} maintain a global index that maps each tool $t$ to an embedding $\phi(d(t))$ of its functional description.
Given a query $q$, the system computes $\phi(q)$ and performs a similarity search over all $N$ embeddings. As a result, the queried tool count is $C(q)=\mathcal{O}(N)$, violating sublinear scaling. Additionally, the centralized index introduces a single point of control and potential failure. Re-indexing after every tool addition or deletion is also non-trivial; batch updates can lead to staleness between index updates and the actual tool set.

\textbf{Context injection.}
Frameworks such as OpenClaw adopt a different strategy by embedding the descriptions $d(t)$ of all tools directly into the LLM's context window. This approach avoids external retrieval components but shifts the scalability burden to the LLM. Inference cost grows quadratically with context length, and the context window itself is inherently bounded. In practice, this limits the approach to at most a few hundred tools. Moreover, any dynamic update to the tool set (adding, removing, or modifying a tool) requires regenerating the entire prompt, incurring noticeable computational overhead.

\textbf{Centralized registries.}
Systems like ANS \cite{ans} and AgentDNS \cite{agentdns} rely on a dedicated registry server that stores all tool metadata. While straightforward to implement, this design places operational control under a single organization, which can create governance challenges \cite{centrial_invisible}. Organizations that prefer not to place full trust in a single central authority may be reluctant to participate. The registry also represents a potential single point of failure \cite{reliable_decentrial}, and deploying and maintaining such a global service requires substantial engineering resources.

\textbf{Over-engineered middleware.}
Proposals such as NANDA \cite{nanda} introduce complex indexing structures (e.g., the Quilt structure) and multi-layer traffic obfuscation to address some of the above concerns. However, these additions bring significant implementation and maintenance complexity, and many of them still rely on a global index at their core, inheriting the same scaling limitations as vector-based retrieval.

What emerges from these paradigms is a recurring pattern as shown in Fig.~\ref{fig:related_work}: each new solution reinvents a global discovery service from scratch above the application layer, and each consequently inherits non-negligible deployment costs, governance friction, and integration overhead. In the next section, we introduce ToolDNS, a framework that reuses and minimally extends the existing DNS to address these limitations. Rather than constructing another global index at a higher layer, ToolDNS maps the discovery problem onto DNS's existing hierarchical namespace, transforming an $\mathcal{O}(N)$ semantic search into an $\mathcal{O}(\log N)$ name resolution process.
\section{ToolDNS System Design}
This section presents our DNS-native framework for AI tool discovery. We begin with a high-level overview of the architecture, then detail the respective designs, including the hierarchical semantic namespace, the query protocol extensions, LLM-augmented semantic pruning, caching mechanisms. We will also discuss practical considerations regarding compatibility, security, and deployment.

\subsection{Overview}
ToolDNS reuses the standard DNS resolution chain, such as root servers, TLD servers, authoritative servers, and recursive resolvers, without modifying their core implementations. As illustrated in Figure ?, the only additions are: 
(i) a reserved top-level domain \path{.tools} that serves as the semantic root for tool discovery; 
(ii) lightweight extensions to the query format (EDNS0 options and partially unfolded domain names) that carry intent and state; 
and (iii) semantic pruning logic inside authoritative servers that are responsible for \path{.tools} subdomains.

The architecture consists of the following logical components:
\begin{itemize}[leftmargin=0.5cm]
    \item \textit{Client (agent)}: Any agent that supports standard DNS queries and EDNS0 can act as a client. The client formulates an intent $q$ and issues a service query (SRV) for a special domain name under \path{.tools}. No custom SDK, protocol adaptation, or central registration is required.
    \item \textit{Recursive resolver}: The resolver performs iterative resolution on behalf of the client. It maintains a cache of delegation records (NS records) and, when necessary, traverses the hierarchy by following referrals. In ToolDNS, the resolver also handles partially unfolded domain names, a construct that encodes the current search position within the domain name itself, and passes the EDNS0 payload unchanged to each authoritative server.
    \item \textit{Root and TLD servers}: The root servers are unchanged; they only need to contain NS records for the \path{.tools} TLD. The TLD servers for \path{.tools} are enhanced with a semantic matching module: given a query with an EDNS0 payload and a partially unfolded name, they return the most relevant subdomains (e.g., \path{weather.tools}, \path{nlp.tools}) instead of a single exact tool matched.
    \item \textit{Intermediate authoritative servers}: These servers manage subdomains deeper in the hierarchy (e.g., \path{history.weather.tools}). Their behavior mirrors that of TLD servers: they receive a partially unfolded name, use the EDNS0 payload to select the top-$K$ matching child subdomains, and return NS records pointing to the next-level authoritative servers.
    \item \textit{Leaf authoritative servers}: These servers directly host tool instances. They store SRV records for fully expanded domain names, along with tool metadata (description, protocol, endpoint). Upon receiving a query that reaches the leaf level, they perform a final semantic match over the local tool list and return the top-$K$ tool endpoints.
\end{itemize}

All these components interoperate with unmodified DNS clients and resolvers that do not support ToolDNS extensions: a legacy resolver will simply treat a \path{.tools} domain name as a normal name and either fail to resolve it (if no A/AAAA record exists) or return an unexpected result. Therefore, ToolDNS is an opt-in enhancement for agents that need semantic discovery, while leaving the global DNS infrastructure untouched.

\subsection{Hierarchical Semantic Namespace}

The core of ToolDNS is a domain name hierarchy that embeds functional semantics and trust information. We design this namespace to support both efficient pruning and decentralized governance.

\subsubsection{Functional hierarchy}

\begin{figure}[t]
\centering
\includegraphics[width=0.45\textwidth]{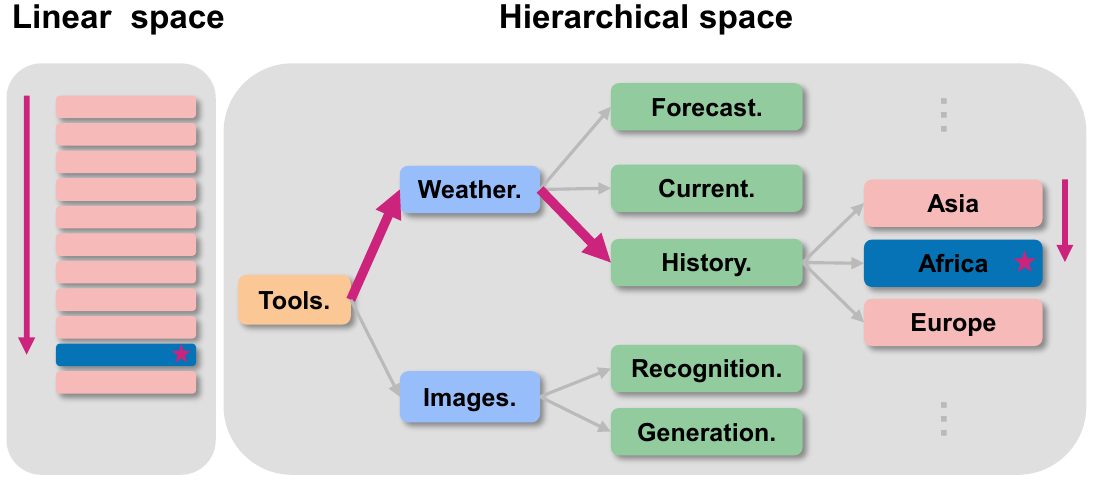}
\caption{Hierarchical domain space architecture of ToolDNS. The diagram illustrates the organization of the tool ecosystem into multiple levels based on their functional semantics.}
\Description{Hierarchical domain space architecture of ToolDNS. The diagram illustrates the organization of the tool ecosystem into multiple levels based on their functional semantics.}
\label{fig:domain_divide}
\end{figure}

As shown in Fig.\ref{fig:domain_divide}, Under the top-level domain \path{.tools}, we construct a tree where each node represents a category or subcategory of tool functionality. The path from the root to a leaf node encodes a progressively refined functional description. For example, a tool that translates English to Chinese and is provided by Alibaba could reside at: \path{english.alibaba.translate.nlp.tools}. Here, each label narrows the scope: \path{nlp} (natural language processing), \path{translate} (translation task), \path{alibaba} (provider), \path{english} (target language). This order places broad categories first {(right) and fine-grained attributes later (left), aligning with top-down pruning.

Formally, let a functional path be a sequence of labels shown as $\ell_1.\ell_2.\cdots.\ell_h.\text{\path{tools}}$, where $h$ is the depth. Each non-leaf node corresponds to a set of subcategories or tool instances. A leaf node is defined as a node that directly hosts tool resource records rather than further subdomains. The hierarchy is not fixed; any authoritative entity can create new subdomains under its delegated zone, allowing the taxonomy to evolve organically.

\begin{rem}[Mapping tools to the hierarchy]
When a tool owner registers a new tool, they identify the most relevant leaf node aligned with the tool's semantics and submit a registration request to the respective subdomain administrator. Upon approval, tool metadata is appended as resource records within that subdomain. Crucially, individual tools do not monopolize a unique domain name; instead, multiple tools sharing the same functional classification coexist under a unified subdomain as separate SRV records.
\end{rem}

\begin{rem}[Search space reduction]
Because each query follows a path from the root downward, the number of tools that must be examined at each step is bounded by the branching factor $\beta$ of the current node. After descending $h$ levels, the candidate set shrinks from the entire universe $N$ to the size of a leaf node's tool list, denoted $S$. This reduces the effective search complexity from $\mathcal{O}(N)$ to $\mathcal{O}(\beta h + S)$, which is $\mathcal{O}(\log N)$ when the tree is balanced.
\end{rem}

\subsubsection{Logical subdomains for decentralized trust}
\begin{figure}[t]
    \centering
    \includegraphics[width=0.45\textwidth]{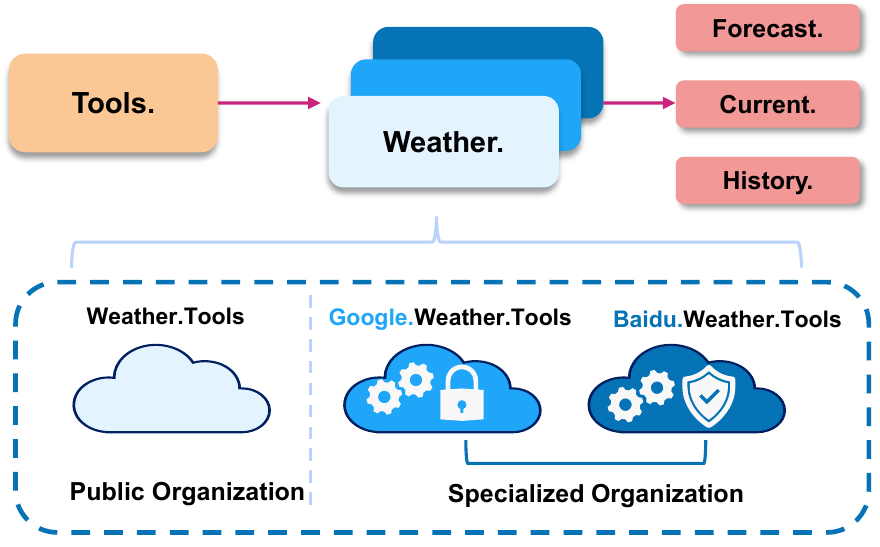}
    \caption{Logic subdomain structure: From the TLD entry point, passing through different organizations (e.g., um/hku; google/baidu) and their functional sub-domains, one can finally lock onto a list of services within a specific niche.}
    \Description{Logic subdomain structure: From the TLD entry point, passing through different organizations (e.g., um/hku; google/baidu) and their functional sub-domains, one can finally lock onto a list of services within a specific niche.}
    \label{fig:logic_domain}
\end{figure}

Traditional DNS leads to a strong coupling between namespace and administrative authority: each subdomain has a single controller. Under such a model, the management authority of a specific subdomain is indivisible, and the publication of its internal information is subject to a single entity. This fails to meet the needs of multiple peer entities (e.g., different universities or companies) that wish to independently maintain tool resources within the same functional category. Developers are forced to choose between ``accepting the management monopoly of a single entity'' and ``fragmenting the namespace by creating separate domains'', making it difficult to simultaneously satisfy global tool interconnection and organization-specific security policies.

{To address this, ToolDNS introduces logical subdomains as shown in Fig.\ref{fig:logic_domain}. The idea is to decouple the functional prefix (e.g., \path{weather.tools}) from the organizational suffix (e.g., ``hku'' in \path{hku.weather.tools} or ``google'' in \path{google.weather.tools}). 
Concretely, the authoritative server for the parent domain \path{.tools} does not solely host \path{weather.tools} as the authoritative server for this subdomain; instead, it simultaneously maintains logical subdomains like \path{hku.weather.tools}, \path{google.weather.tools}, and \path{noaa.weather.tools}, delegating them to independent authoritative servers operated by each organization. From the perspective of the resolution process, a query for the domain \path{_service._proto._tools} with a task description matching "weather" in EDNS0 will cause the \path{.tools} server to return multiple NS records, including one for \path{weather.tools}, and potentially others for logical subdomains like \path{google.weather.tools} or \path{noaa.weather.tools}.

An agent that wishes to enforce a trust policy can simply restrict its resolution to specific organizational subdomains. For example, an agent that only trusts tools endorsed by HKU can set its initial query name to \path{_service._proto._hku.weather.tools} instead of the generic \path{_weather.tools}. This capability requires no additional trust anchor or public key infrastructure (PKI); it leverages the existing delegation chain (the parent domain's authority guarantees that \path{hku.weather.tools} is indeed managed by HKU).

We distinguish two classes of subdomains:
\begin{enumerate}[leftmargin=0.5cm]
    \item \textit{Public common class}: These subdomains typically consist of ``official'' followed immediately by a functional prefix, with ``official'' usually being hidden (e.g., \path{official.weather.tools} $\rightarrow$ \path{weather.tools}). They serve as open entry points and may be managed by a community body or the \path{.tools} registry. Their purpose is to provide a neutral discovery path for agents that do not have specific trust requirements.
    \item \textit{Certified trust class}: These are subdomains that include an organizational identifier (e.g., \path{hku.weather.tools}). The identifier is typically placed immediately to the left of the functional prefix. By resolving through such a subdomain, an agent receives tools that are directly endorsed and managed by the named organization. This design enables verifiable, accountable service discovery without centralization.
\end{enumerate}

The two classes can coexist seamlessly: the same functional prefix can have both a public entry point and multiple certified subdomains. An agent may query the public entry first and, if unsatisfied, restrict to a trusted subdomain, or it may query only certified subdomains from the start.

\subsection{Query Protocol and Semantic Encoding}

Standard DNS was designed for exact name lookup: the client must know the full domain name before issuing a query. In tool discovery, however, the client knows only an intent. To bridge this gap, we introduce several lightweight protocol extensions that remain fully compliant with DNS specifications.

\subsubsection{Protocol support}
The current AI tool ecosystem is characterized by protocol fragmentation. Mainstream standards such as MCP, A2A, Skill, and REST-ful API, are maintained by disparate developer communities, leading to a discovery process often restricted to specific frameworks. ToolDNS does not aim to define a new universal protocol or add an adaptation layer; instead, it directly encodes the protocol specifications of tools into the query requests. By utilizing the \path{_service} and \path{_protocol} labels of DNS SRV records within the domain name structure, ToolDNS establishes a globally universal tool index directory. As long as a tool provides its protocol attributes to DNS system according to the standard field of SRV record (e.g., \path{service = mcp and protocol=tcp}), it can be discovered by any ToolDNS-compatible agent, achieving unification at the retrieval level.

\subsubsection{Partially unfolded domain names}
Traditional DNS requires a fully qualified domain name (FQDN) to retrieve specific resource records. In AI tool discovery, the target service's domain name is unknown in advance because the discovery process is driven by dynamic intent rather than static identifiers. To address this while maintaining full compatibility, we introduce the concept of partially unfolded domain names.

We define two forms of domain names used in ToolDNS queries:
\begin{itemize}[leftmargin=0.5cm]
    \item \textit{Fully expanded domain name}: This follows the standard SRV record format: \path{_service._proto.domain.} (with no leading underscore before \path{domain}). It indicates that the resolution has reached a leaf node, and the \path{domain} part can be resolved to an IP address via A/AAAA records. For example, \path{_mcp._tcp.api.history.weather.tools.} is a fully expanded name.
    \item \textit{Partially expanded domain name}: This extends the SRV format by inserting an extra underscore before the \path{domain} part: \path{_service._proto._.domain.}. The underscore acts as a search cursor: it marks that the path is still under construction and that the resolver should continue traversing deeper. The cursor is a placeholder that will be replaced by the next matched subdomain label.
\end{itemize}

The resolution process evolves a partially unfolded name from right to left, similar to unrolling a scroll. Initially, the resolver knows only the service type and protocol (e.g., \path{_mcp._tcp}). It starts with the minimal cursor \path{_tools.}, producing \path{_mcp._tcp._tools.}. When it receives NS records for a subdomain (say \path{weather.tools}), it inserts that label between the cursor and the \path{domain} part, yielding \path{_mcp._tcp._weather.tools.}. When the next step matches \path{history.weather.tools}, the name becomes \path{_mcp._tcp._history.weather.tools.}. Finally, upon reaching a leaf node, the resolver removes the leading underscore, obtaining the fully expanded name \path{_mcp._tcp.history.weather.tools.} and then queries for its SRV records.

This design ensures that every step of the discovery is expressed as a standard DNS query name, making the entire process compatible with unmodified recursive resolvers (as long as they forward EDNS0 options). The cursor mechanism encodes state without requiring the resolver to maintain per-query state machines; the state is carried in the name itself.

\subsubsection{EDNS0 semantic payload}
While partially unfolded names capture the where of the search (the current position in the hierarchy), they cannot carry the what (i.e., the user's natural language intent) because DNS domains are limited to $253$ bytes. To transmit the intent, we use the EDNS0 mechanism, which allows a DNS query to include additional option data in its request packet.

\begin{figure}[t]
    \centering
    \includegraphics[width=0.48\textwidth]{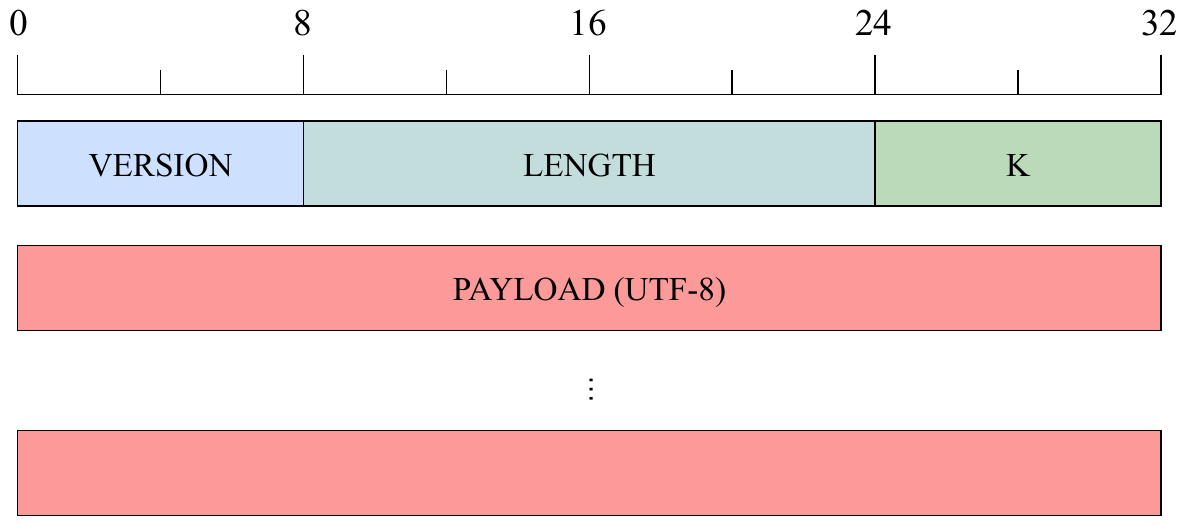}
    \caption{Packet format (payload starts at bit 32).}
    \Description{Structure of the packet format. The header occupies the first 32 bits, divided into three fields: VERSION (8 bits), LENGTH (16 bits), and K (8 bits). Below the header, the payload is shown as a variable-length UTF-8 encoded field that may span multiple rows, indicated by a continuation box and ellipsis.}
    \label{fig:EDNS0_structure}
\end{figure}

We define a new EDNS0 packet structure with the binary layout shown in Figure \ref{fig:EDNS0_structure}. The option data consists of:
\begin{itemize}[leftmargin=0.4cm]
    \item \textit{Version ($8$ bits)}: Currently set to \path{0x00}. Future revisions of the payload format can increment this field while using the same option code, ensuring backward compatibility.
    \item \textit{Length ($16$ bits)}: The length in bytes of the following Payload field (not including Version, Length, or $K$). This allows payloads up to $65,531$ bytes, sufficient for complex natural language queries.
    \item \textit{$K$ ($8$ bits)}: The number (unsigned integer) of top results requested at each semantic pruning step. A value of $K=0$ is reserved for special use (e.g., cache warming, as discussed in Section \ref{sec:cache}).
    \item \textit{Payload (variable)}: A UTF-8 encoded string containing the agent's intent. For version 0, this is plain text; future versions may support compressed or simple structured formats (e.g., CBOR).
\end{itemize}

An agent constructing a query sets the EDNS0 option with its intent and the desired $K$. The recursive resolver must preserve this option exactly when forwarding queries to authoritative servers. Servers that do not recognize the option ignore it and return a normal (non-semantic) response, causing the resolution to fail gracefully. This is acceptable because only ToolDNS-aware servers can perform semantic pruning.

\begin{rem}[An example]
   For the intent ``fetch historical weather data for Hong Kong'', the EDNS0 payload might contain the string ``historical weather Hong Kong''. The recursive resolver includes this in every query it sends during iterative resolution. At the \path{weather.tools} server, the payload is compared against the semantic summaries of its child subdomains (e.g., \path{history.weather.tools}, \path{forecast.weather.tools}), and the most relevant subdomains are returned. 
\end{rem}

\subsection{Iterative Resolution Algorithm}
\begin{figure}
    \centering
    \includegraphics[width=0.5\textwidth]{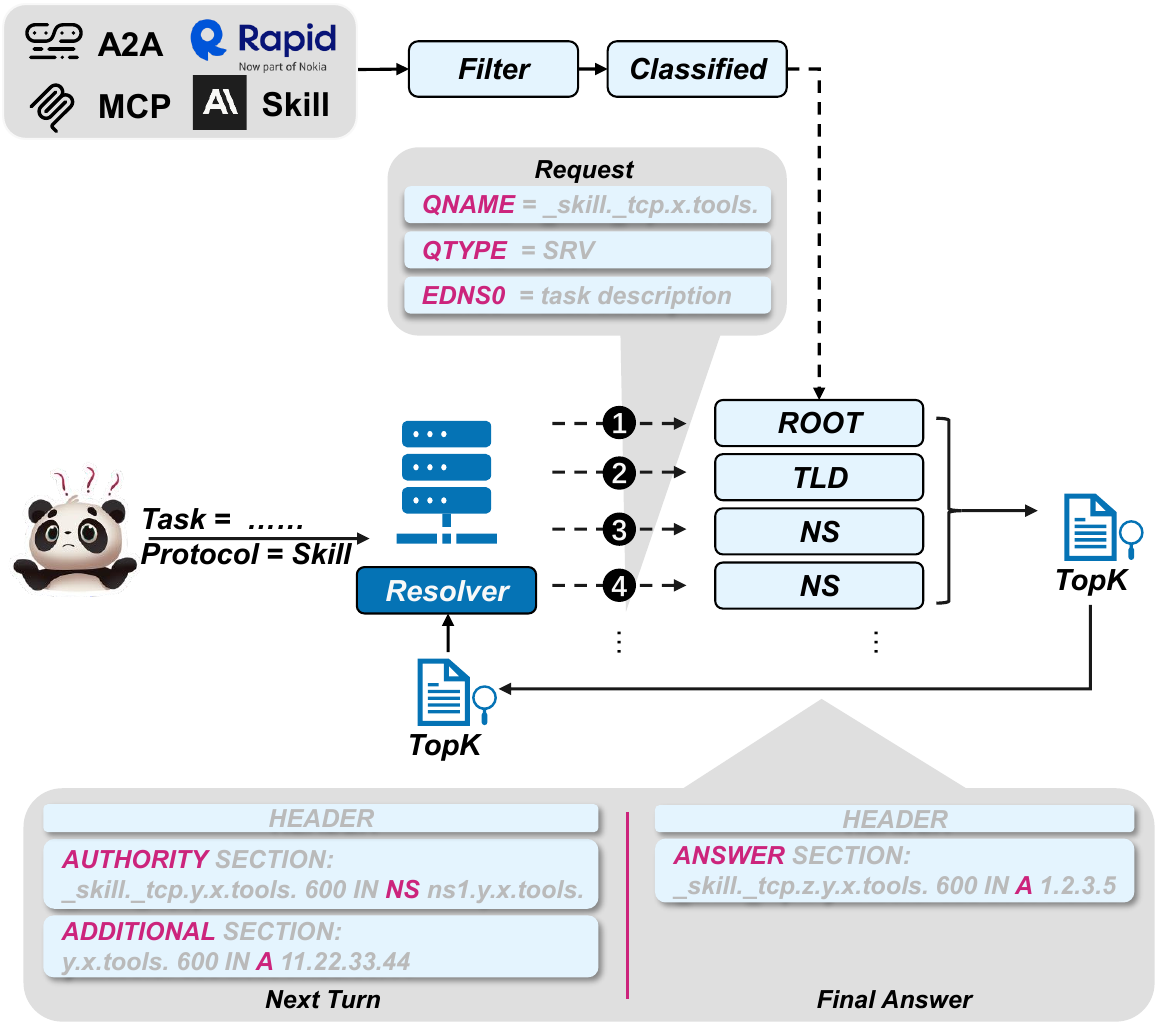}
    \caption{The comprehensive architecture and mechanism of ToolDNS. It illustrates the server structure, dataset construction, and the iterative discovery workflow, highlighting the request/response packet structures.}
    \label{fig:ToolDNS_framework}
\end{figure}
With the namespace and query extensions in place, we now present the complete discovery algorithm as executed by the recursive resolver. Algorithm \ref{alg:discovery} summarizes the procedure shown in Fig.\ref{fig:ToolDNS_framework}.

\begin{algorithm}[t]
    \SetAlgoLined
    \SetKwInOut{Input}{Input}
    \SetKwInOut{Output}{Output}
    
    \Input{Service type $s$ (e.g., ``mcp''), protocol $p$ (e.g., ``tcp''), intent string $I$, integer $K$}
    \Output{List of $(IP\ address,\ port,\ protocol)$ for top-$K$ tools}
    
    \BlankLine
    Initialize $\textit{domain} \gets \texttt{\_}\/s\texttt{.\_}\/p\texttt{.\_tools.}$\;
    \While{true}{
        Send SRV query for $\textit{domain}$ with EDNS0$(K, I)$\;
        Receive response $R$\;
        \If{$R$ contains answer records (SRV)}{
            \ForEach{SRV record in $R$}{
                Resolve target name via A/AAAA query\;
            }
            \Return list of endpoints\;
        }
        \ElseIf{$R$ contains authority records (NS)}{
            Let $\textit{subdomains} \gets$ extract {Authoritative Domain of NS from} $R$ (max $K$ entries)\;
            Choose one subdomain $d$ from $\textit{subdomains}$\;
            \tcp{Insert $d$ before the cursor}
            {Replace cursor ``\_'' before $\textit{domain}$ with $\_d$\texttt{.}\;}
            \tcp{Continue to next iteration}
        }
        \Else{
            \Return empty list \tcp{no match or error}
        }
    }
    \caption{ToolDNS discovery at the recursive resolver.}
    \label{alg:discovery}
\end{algorithm}

The algorithm begins with the minimal partially unfolded name that includes only the service and protocol (line 1). The cursor is \path{_tools.}, the root of the semantic namespace. In each iteration, the resolver sends an SRV query carrying the EDNS0 payload (line 3). The authoritative server at the current level responds either with NS records (if the current node is non-leaf) or with SRV records (if it is a leaf). When NS records are returned, the resolver selects one subdomain (typically the first among the top-$K$; the agent may choose to try multiple) and updates the query name by replacing the cursor with that subdomain label (line 14). The loop continues until SRV records are obtained.

As an example, Suppose an agent wants an MCP tool for historical weather. The resolver starts with \path{_mcp._tcp._tools.}. The \path{.tools} TLD server returns NS records for \path{weather.tools} and \path{nlp.tools} (assuming $K=2$). The resolver picks \path{weather.tools} and updates the name to \path{_mcp._tcp._weather.tools.}. The \path{weather.tools} server returns NS records for \path{history.weather.tools} and \path{forecast.weather.tools}. The resolver chooses \path{history.weather.tools} and updates to \path{_mcp._tcp._history.weather.tools.}. The server for \path{history.weather.tools} is a leaf; it returns SRV records for tools like \path{api.history.weather.tools}. The resolver then removes the underscore and resolves \path{api.history.weather.tools} to an IP address via a standard A query.

\subsection{LLM-augmented Semantic Pruning at Authoritative Servers}

The efficiency of ToolDNS hinges on the ability of each authoritative server to return the most relevant subdomains (or tools) given an intent. We implement this semantic pruning using a lightweight, LLM-augmented matching module. The module is stateless and does not require a global index; each server operates only on its local zone data.

\subsubsection{Non-leaf servers (TLD and intermediate)}
Each non-leaf server maintains a list of its child subdomains. For each child, it stores a short semantic summary, typically a few keywords or a sentence extracted from the child's description. The summaries can be generated offline using an LLM or even manually defined by the zone administrator. Upon receiving a query with intent $I$ and parameter $K$, the server computes a relevance score between $I$ and each child's summary using a fast similarity function. Options include: 
\begin{itemize}[leftmargin=0.5cm]
    \item \textit{Embedding-based cosine similarity}: Pre-compute embeddings for each child summary; at query time, embed $I$ and compute dot products. This yields high accuracy but requires an embedding model.
    \item \textit{Keyword matching}: Use TF-IDF or BM25 on the summaries. Faster but less accurate.
    \item \textit{Small LLM scoring}: For maximum adaptability, the server can invoke a tiny LLM (e.g., a sub-$10$B active-parameters model) to score the top few candidates.
\end{itemize}

The server then returns the NS records of the $K$ children with the highest scores. The choice of $K$ controls the trade-off between exploration (larger $K$ reduces the chance of pruning a relevant branch) and efficiency (smaller $K$ reduces subsequent work). In our evaluation, we use small LLM scoring, and set $K=1$ for the highest pruning ratio and show that accuracy remains high due to the hierarchical semantics.

\subsubsection{Leaf servers}
A leaf server directly hosts tool instances, each with a functional description $d(t)$ (the tool's natural language documentation). The matching procedure is similar: the server computes relevance between $I$ and each $d(t)$ and returns the top-$K$ tools in the form of SRV records. Each SRV record's target name is a fully expanded domain name (e.g., \path{api.history.weather.tools}), and the port and protocol are encoded in the SRV fields.

\begin{rem}[Global-index-free operation]
 A crucial advantage of this design is that no global index needs to be rebuilt when tools are added, removed, or updated. Adding a new tool simply requires appending an SRV record to the leaf server's zone file (or updating a database backend). The next query will immediately reflect the change because the matching is performed online. This contrasts with vector-based retrieval systems that require periodic re-indexing of the entire tool set.   
\end{rem}

\begin{rem}[LLM deployment considerations]
   While we call this LLM-augmented, the actual output of LLM is only several category names for subdomains, which leads to low latency for LLM inference and can be run on a single consumer GPU, with negligible latency (a few milliseconds). For very high-throughput servers, one can pre-compute embeddings for child summaries and use fast approximate nearest neighbor search. The recursive resolver does not need to know which method is used; it only sees the resulting NS/SRV records. 
\end{rem}
\section{Governance And Practical Considerations}
Beyond the core resolution and pruning logic, several practical factors are essential for the deployment of ToolDNS in real world. This section introduces the system's approach to caching efficiency, security assurance, and compatibility with future demands.
\subsection{Caching and Performance Optimization}\label{sec:cache}
\begin{figure}[t]
    \centering
    \includegraphics[width=0.45\textwidth]{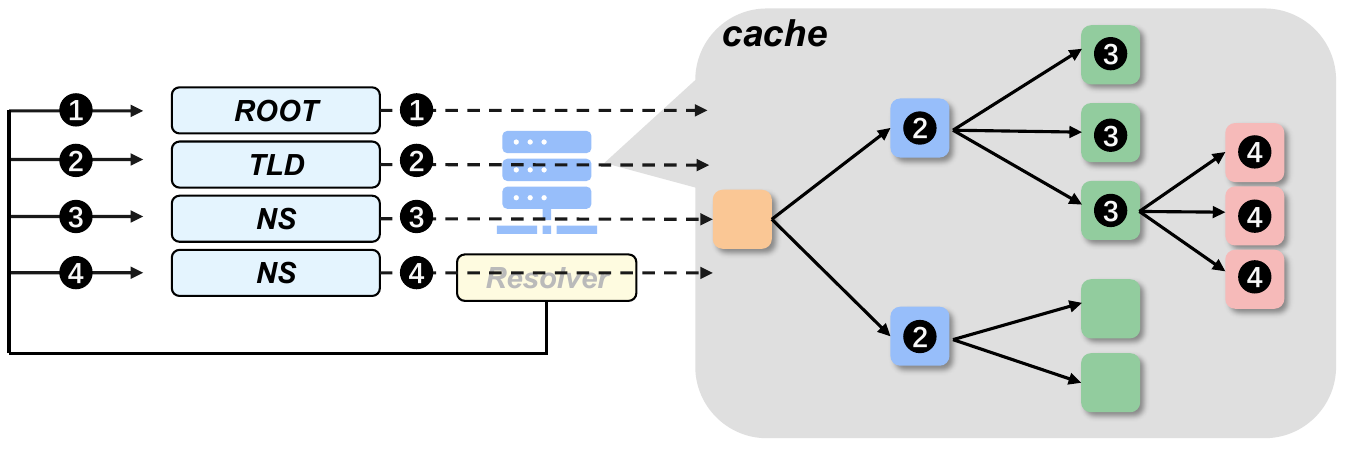}
    \caption{Caching Mechanism of Recursive Resolver. This mechanism demonstrates how a recursive resolver synchronizes the complete directory tree structure through initial queries, thereby transforming remote cross-network queries into low-latency local semantic filtering and lookups.}
    \Description{Caching Mechanism of Recursive Resolver. This mechanism demonstrates how a recursive resolver synchronizes the complete directory tree structure through initial queries, thereby transforming remote cross-network queries into low-latency local semantic filtering and lookups.}
    \label{fig:cache}
\end{figure}

Our DNS-based approach offers significant potential for further efficiency optimization, particularly through the exploitation of native DNS caching mechanisms to minimize round trip times. ToolDNS adds a proactive cache warming strategy as shown in Fig.~\ref{fig:cache}.

The recursive resolver caches two types of information:
\begin{enumerate}[leftmargin=0.5cm]
    \item \textit{Delegation (NS) records for non-leaf nodes}: These records have a long Time-To-Live (TTL), typically hours or days, because the hierarchical structure (e.g., which subdomains exist under \path{weather.tools}) changes infrequently. Once cached, the resolver can skip one or more RTTs when answering subsequent queries.
    \item \textit{Tool instance (SRV) records for leaf nodes}: These have a shorter TTL (minutes) to reflect the dynamic nature of tool availability and endpoints.
\end{enumerate}

The resolver's cache stores a tree of subdomain names, each associated with its NS records and optional semantic summaries. When a new query arrives, the resolver can use the cache in two manners:
\begin{itemize}[leftmargin=0.5cm]
    \item \textit{Server address search}: If the resolver knows the part of the domain, the resolver first looks for the longest suffix match in the cache that corresponds to a fully expanded or partially unfolded name. If the resolver ends at a non-leaf node of the tree, it can continue to send requests to real authoritative DNS server for the next subdomain. If the entire path to a leaf is cached, the resolver can directly query the leaf server without contacting intermediate servers.
    \item \textit{Query mock}: If the resolver does not know any information about the target domain, the resolver does not send request to a DNS server, instead, it skips the request and gets the response directly from the cache tree. Then the resolver can run a semantic match for top-$K$ subdomains until it runs beyond the tree. Then it continues to send request to the real authoritative DNS server for next subdomain.
\end{itemize}

To further reduce cold-start latency, we introduce a special mode: when $K=0$, the authoritative server is requested to return all child subdomains (or all tools) at the current level, rather than only the top-$K$. The resolver can issue such a query during off-peak hours or when it first joins the network, thereby populating its cache with the complete directory structure. After warming, most queries will hit the cache within the resolver and only need to contact the leaf server for the final tool list.

In the warm-cache case, the end-to-end discovery latency consists of: 
(i) one RTT to the leaf server (if the leaf's SRV records are not cached) plus 
(ii) one RTT to resolve the final A/AAAA record. 
This is comparable to the latency of a centralized registry (which also requires at least one RTT for the lookup and one for the endpoint resolution). 
In the worst-cold case, the number of RTTs equals the depth of the hierarchy (typically $2$-$4$). Our experiments in Section \ref{sec:exp} show that this overhead is acceptable given the scalability and decentralization benefits.

\subsection{Practical Considerations}
We conclude the system design with a discussion of compatibility, security, and deployment.

\textbf{Compatibility with existing DNS.}
ToolDNS does not modify any DNS protocol messages except for adding a new EDNS0 option, which is explicitly allowed by the standard. Unmodified recursive resolvers will ignore the option and forward queries as usual; they will not perform semantic pruning, so the query will likely fail to resolve because \path{.tools} authoritative servers expect the option. However, an agent can fall back to a conventional discovery method if the EDNS0 option is not supported. More importantly, ToolDNS does not require changes to the root servers or to the vast majority of DNS infrastructure; only the authoritative servers for \path{.tools} and its subdomains need to be enhanced.

\textbf{Security and trust.}
The logical subdomain mechanism inherits the security properties of DNS delegation. If an agent trusts a particular organization (e.g., HKU), it can directly query \path{_service._proto._hku.weather.tools}, and the response will come from HKU's authoritative servers. The authenticity of the NS delegation can be verified with DNS security extensions (DNSSEC) if the \path{.tools} zone and the organization's zone are signed. We do not require DNSSEC for correct operation, but it can be layered on top for added integrity. The EDNS0 payload is sent in clear text; for privacy-sensitive intents, agents should use DNS over TLS (DoT\cite{DoT}) or DNS over HTTPS (DoH\cite{DoH}) between the resolver and the authoritative servers.

\textbf{Deployment roadmap.}
Deploying ToolDNS at global scale requires only two coordinated actions: (i) IANA or a suitable authority delegates the \path{.tools} TLD and sets up TLD servers with semantic pruning; (ii) organizations that wish to publish tools register their logical subdomains under the appropriate functional prefixes. No changes to client operating systems or network stacks are needed, as standard DNS libraries already support SRV queries and EDNS0. This low barrier to entry is a key advantage over alternative proposals that require new infrastructure.

\subsection{Forward Compatibility}
ToolDNS assumes that tool functionality can be captured by a hierarchical taxonomy. For highly niche tools domain, the taxonomy may be incomplete. However, in that case, ToolDNS can degrade into a proprietary or centralized solution, proprietary implementations within the subdomain can be specially optimized based on actual conditions. In this scenario, the ToolDNS system directs only relevant queries to that implementation and diverts unrelated ones to other subdomains in advance, this also demonstrates the architectural flexibility and forward compatibility of ToolDNS.

While the current ToolDNS utilizes plain text intents to maximize parsing throughput and minimize computational overhead, future iterations could incorporate structured query schemas (e.g., ProtoBuf encoding) within EDNS0 options. This evolution could enable constraint aware matching, allowing clients to specify required parameter types or constraints directly during the discovery phase. By introducing structured payloads, DNS will evolve from simple intent resolution into a parameter aware discovery service within the tool discovery ecosystem. 

Finally, while the reliance on a dedicated \path{.tools} TLD requires coordination with DNS root governance and the establishment of various sub-domains, these are one-time initialization efforts. Furthermore, this model distributes the maintenance costs across specialized domains, ensuring that the one-time workload within each domain remains extremely low.
\section{Experiments}\label{sec:exp}
The preceding sections have established ToolDNS as a hierarchical, DNS-native framework for AI tool discovery and analyzed its theoretical scaling properties. In this section, we empirically validate these claims through a comprehensive experimental evaluation. Our assessment focuses on two interdependent dimensions that jointly determine the practical utility of any discovery system: {effectiveness} (how accurately the system retrieves relevant tools given natural language intents) and {efficiency} (how the search space grows with the tool ecosystem and how the DNS-native approach alleviates network overhead).

\subsection{Heterogeneous Dataset Construction}
A fundamental obstacle to evaluating AI tool discovery across emerging protocols is the absence of a unified, large-scale benchmark. While prior work, such as ToolBench, provides extensive coverage of traditional RESTful APIs, there is a conspicuous lack of evaluation data for newer, agent-centric standards like OpenClaw Skills, the Agent-to-Agent (A2A) protocol, and Model Context Protocol (MCP) tools. Furthermore, existing MCP collections (e.g., MCPZoo \cite{mcpzoo}) often lack standardized functional descriptions, which are indispensable for intent-based retrieval. To bridge this gap and rigorously assess the universality of ToolDNS, we curated an integrated dataset from the following heterogeneous sources:
\begin{itemize}[leftmargin=0.5cm]
    \item \textbf{RESTful API}: Sourced from the G1 task set of the ToolBench benchmark, representing traditional Web API specifications;
    \item \textbf{MCP Tools}: Collected from the MCP toolset on MCPZoo\cite{mcpzoo}, representing the emerging ecosystem of model context protocols;
    \item \textbf{OpenClaw Skills}: Based on the community-maintained Awesome OpenClaw Skills list, representing skill invocation standards;
    \item \textbf{A2A}: Based on the official A2A protocol examples released by Google, representing communication specifications between agents.
\end{itemize}

To transform this raw collection into a clean, hierarchically organized benchmark with query-intent pairs, we executed a multi-stage data processing pipeline using Qwen3-30B-A3B-Instruct-2507 \cite{qwen3technicalreport} (ereinafter referred to as Qwen) and DeepSeek. The procedure, summarized below, ensures both the quality of the data and the semantic coherence of the taxonomy required by ToolDNS:

    \textbf{Step 1. Functional summarization}. We utilized Qwen to generate summaries for the descriptions of all raw tools to facilitate subsequent processing.
    
    \textbf{Step 2. Low-quality data removal}. We employed Qwen to filter out invalid tools with vague functional descriptions, lack of practical significance, or missing documentation. As shown in Table~\ref{tab:data_cleaning_stats}, this step removed 3,730 samples, accounting for $6.82\%$ of the raw data.
    
    \textbf{Step 3. Taxonomy construction}. With the assistance of models such as DeepSeek and manual labor, we extracted top-level categories from a macro perspective to establish an initial directory topology.
    
    \textbf{Step 4. Automated coarse classification}. We used Qwen to preliminarily classify all tools according to the categories extracted in the previous step, removing those that could not be classified. As shown in Table~\ref{tab:data_cleaning_stats}, This step eliminated $7,359$ samples, bringing the cumulative removal rate to $13.45\%$.
    
    \textbf{Step 5. Sub-category refinement}. For data within each major category, we performed sub-category division with the assistance of models like DeepSeek and manual labor.
    
    \textbf{Step 6. Automated fine classification and refinement}. We again used Qwen to assign sub-categories to data within each major category and removed unclassifiable items. As shown in Table~\ref{tab:data_cleaning_stats}, This step removed $9,923$ samples, resulting in a final cumulative removal rate of $38.41\%$.
    
    \textbf{Step 7. Evaluation baseline construction}. Following the methodology of ToolLLM, we used Qwen to generate task queries for each tool. These ``intent-tool'' pairs serve as the ground truth to test the retrieval effectiveness of ToolDNS.

\begin{table}[t]
  \centering
\setlength{\tabcolsep}{3.1mm}
  \caption{Statistical Summary of the Data Cleaning Process}
  \label{tab:data_cleaning_stats}
  \begin{tabular}{l l r r r}
    \toprule
    \textbf{Step} & \textbf{Removed} & \textbf{Remaining} & \textbf{Rate (\%)} \\
    \midrule
    Raw Data & --- & $54700$ & 0.00 \\
    \midrule
    Step 2 & 3730 & 50970 & 6.82 \\
    Step 4 & 7359 & 43611 & 13.45 \\
    Step 6 & 9923 & 100 & 18.14\\
    \midrule
    \textbf{Total} & \textbf{$21012$} & \textbf{$33688$} & \textbf{38.41\%} \\
    \bottomrule
  \end{tabular}
\end{table}

\begin{table*}[t]
    \centering
    \caption{Comparative Analysis of Network Overhead and Query Efficiency over IP Protocols}
    \label{tab:protocol_cost}
    \setlength{\tabcolsep}{4.5pt}
    \begin{tabular}{lccccccccc}
        \toprule
        & \multicolumn{4}{c}{\textbf{Request}} & \multicolumn{4}{c}{\textbf{Response}} \\
        \cmidrule(lr){2-5} \cmidrule(lr){6-9}
        \textbf{Scheme} & Total (MB) & Packets & \makecell{Bytes (KB)/\\Query} & \makecell{Packets/\\Query} & Total (MB) & Packets & \makecell{Bytes (KB)/\\Query} & \makecell{Packets/\\Query} & \makecell{\textbf{Delay (ms)}/\\Query}\\
        \midrule
        AgentDNS & 15.69 & 145,107 & 1.16 & 10.77 & 24.10 & 133,020 & 1.79 & 9.87 & 2035.51\\
        ANS      & 14.66 & 145,108 & 1.09 & 10.77 & 18.96 & 133,022 & 1.41 & 9.87 & 2042.68\\
        \textbf{ToolDNS} & \textbf{9.70} & \textbf{40,198} & \textbf{0.65} & \textbf{2.98} & \textbf{4.40} & \textbf{40,198} & \textbf{0.33} & \textbf{2.98} & \textbf{5.62}\\
        \bottomrule
    \end{tabular}
\end{table*}

\subsection{End-to-End Query Hit Rate Comparison}
\label{subsec:hit_rate}
\begin{figure}[t]
    \centering
    \includegraphics[width=0.45\textwidth]{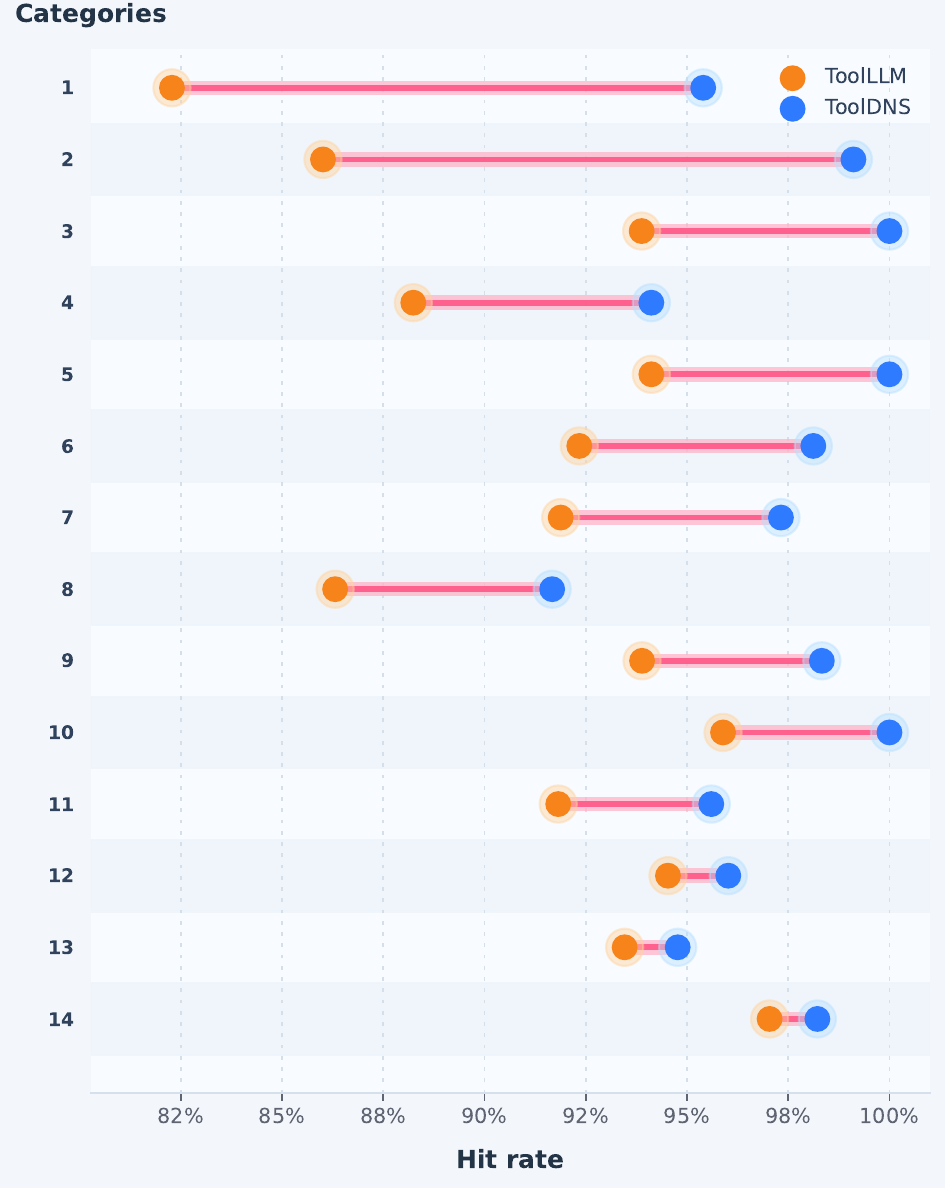}
    \caption{Retrieval accuracy comparison: ToolDNS versus the ToolLLM baseline. (A full enumeration of categories is provided in Appendix~\ref{app:tool_categories}.}
    \Description{Retrieval accuracy comparison: ToolDNS versus the ToolLLM baseline.}
    \label{fig:acc}
\end{figure}

With a robust, hierarchically structured benchmark in place, we first seek to answer the most fundamental question: does the pruning inherent in hierarchical discovery preserve retrieval accuracy? While reducing the search space is essential for scalability, such gains must not come at the cost of delivering irrelevant or incorrect tools to the agent. This first experiment evaluates the fidelity of ToolDNS by measuring its hit rate against a state-of-the-art flat retrieval baseline.

For any given tool $T$, we use its functional summary as the indexing feature and its associated task description as the query input. A retrieval is counted as a hit if the returned tool belongs to the same semantic sub-category as $T$.
To ensure a fair comparison, we partitioned the dataset into a training set (60\%) and a test set (40\%) using a stratified split from the {sklearn} library. We retrained the ToolLLM retriever on the training set, strictly adhering to the methodology described in the original paper \cite{toolllm}. Both ToolDNS and the ToolLLM baseline were then evaluated on the identical held-out test set. The System prompt used in this experiment is provided as ``Agent Role of TLD server'' and ``Agent Role of NS server'' in Appendix.\ref{app:prompts}

As illustrated in Figure~\ref{fig:acc}, ToolDNS achieves a superior overall hit rate across the evaluated domains. This indicates that the hierarchical partitioning employed by ToolDNS does more than just prune the search space; it actively suppresses retrieval noise by constraining the candidate pool to semantically coherent clusters. These results confirms the high quality of both the constructed dataset and the hierarchical taxonomy, and validate that our scheme improves the retrieval robustness of flat vector search.

\subsection{Computational Complexity and Search Space Reduction}
Given that ToolDNS maintains high retrieval accuracy, the next natural question concerns the magnitude of its computational advantage.
Existing tool discovery mechanisms are confronted with the challenge of search space explosion. As the number of tools scales to tens of thousands or beyond, the traditional global scan approach suffers from severe inefficiency and prohibitive computational costs. In this experiment, we simulate tool repositories of varying scales to evaluate the performance of ToolDNS against traditional flat retrieval schemes (e.g., ToolLLM), specifically focusing on the magnitude of the search space required to be explored during a single task-matching process.

We compare the size of the search space for a single query under both schemes as the number of tools increased from $10^3$ to $3.4 \times 10^4$. For ToolDNS, we configured the system to return the Top-1 subdomain at each layer, with a total of 2 layers of classification directories. The experimental results are shown in Fig.~\ref{fig:compute_complexity}. 

At a scale of $2.53 \times 10^4$ tools, a single-layer ToolDNS classification reduces the active search space to $4,888.10$ tools (on average), achieving an $80.65\%$ reduction compared to the exhaustive scan of ToolLLM. With two classification layers, the search space shrinks dramatically to $1,197.12$ tools, representing a $95.26\%$ reduction. The advantage becomes even more stark at $3.03 \times 10^4$ tools, where the two-layer hierarchy reduces the search space by over $99.99\%$ compared to the full context injection approach of OpenClaw. It is important to note that our dataset taxonomy is only refined to two levels; thus, these observed reductions are still far from the theoretical $\mathcal{O}(\log N)$ upper bound that deeper, well-balanced hierarchies could achieve.

\begin{figure}[t]
    \centering
    \includegraphics[width=0.45\textwidth]{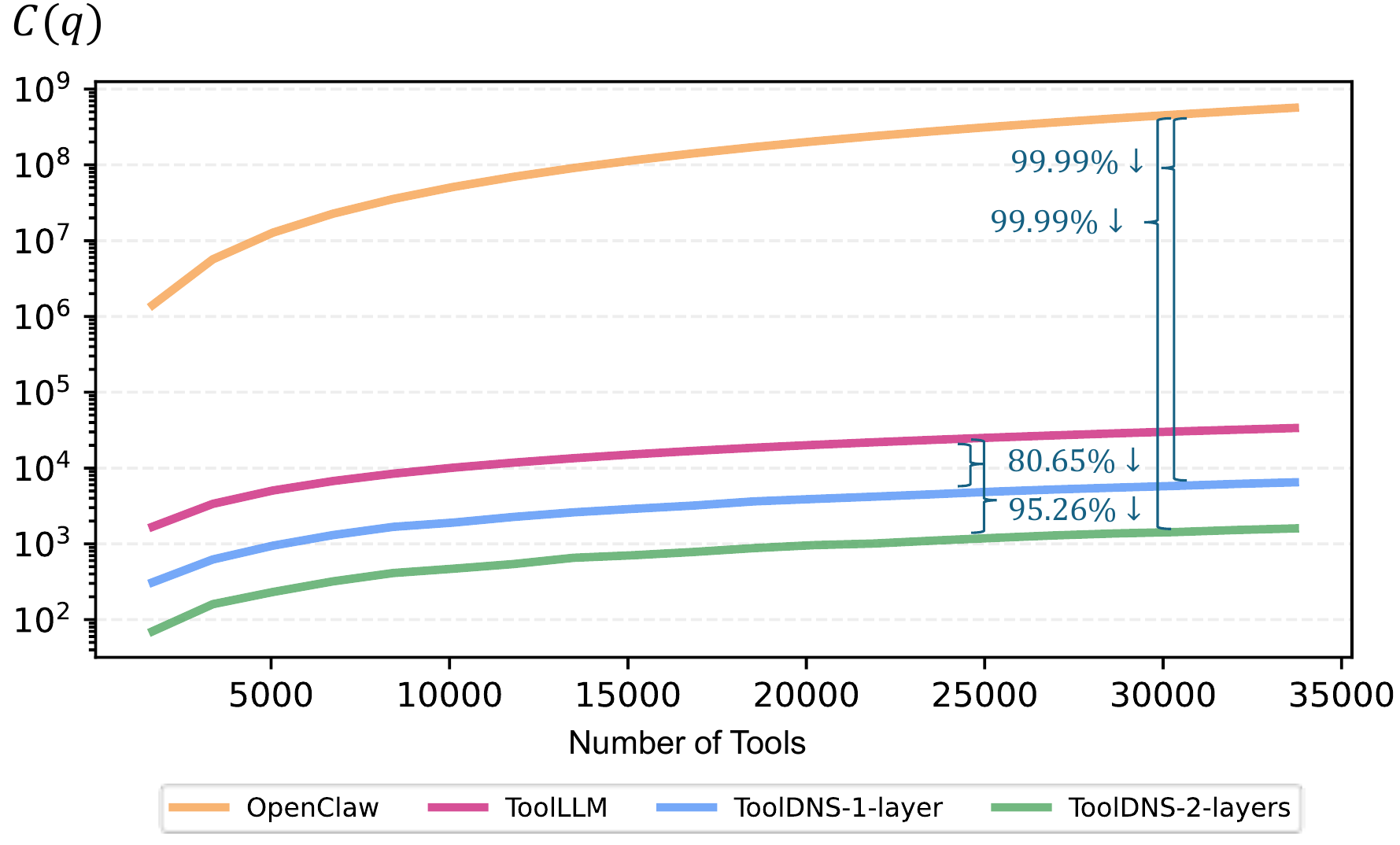}
    \caption{Scalability of search space size. The comparison between ToolDNS and baseline flat retrieval schemes demonstrates that hierarchical classification effectively curbs search space growth as the tool ecosystem expands.}
    \Description{Scalability of search space size. The comparison between ToolDNS and baseline flat retrieval schemes demonstrates that hierarchical classification effectively curbs search space growth as the tool ecosystem expands.}
    \label{fig:compute_complexity}
\end{figure}

\subsection{Comparative Evaluation of Network Efficiency}

The previous experiment established that ToolDNS drastically curtails the number of tools examined per query. However, even a computationally efficient discovery process can be hindered by verbose communication protocols. In this experiment, we measure the actual network traffic incurred during the discovery process to evaluate the lightweight nature of the DNS-native approach.

We conduct a large scale experiment consisting of more than $13,000$ tool discovery queries (using the test data described in Section~\ref{subsec:hit_rate}) and measure traffic volume and packet overhead, including total upstream/downstream bytes and packet counts. We compare ToolDNS against two representative HTTP-based baselines (AgentDNS and ANS).

To avoid biasing the comparison against ToolDNS, we adopt a conservative experimental setup for the HTTP baselines. Specifically, we implement simplified versions according to the data structures described in their original papers and reserve only the core fields required for the tool discovery task. This minimizes protocol overhead on the baseline side and intentionally biases the comparison in their favor. As a result, the measured performance can be regarded as a lower bound on the network cost of HTTP-based approaches. In contrast, ToolDNS operates with its standard message format without additional optimization.

As shown in Table~\ref{tab:protocol_cost}, ToolDNS significantly reduces both upstream and downstream data transfer compared to the baselines. This reduction stems from the lightweight DNS message format, which avoids the verbose headers and structured payloads required by HTTP-based systems. In contrast, the baselines incur substantial overhead from HTTP headers, TCP handshakes, and request/response bodies in JSON format. Consequently, ToolDNS achieves a markedly lower total bandwidth consumption across all queries. Beyond total volume, the HTTP-based baselines incur significant packet overhead from mandatory TCP handshakes and connection management. ToolDNS, leveraging the connectionless nature of UDP, completes most discovery tasks within a single query and response cycle, minimizing the total IP packet count.

These measurements confirm that ToolDNS benefits from a fundamentally more efficient communication protocol, compounding the computational search space reductions demonstrated earlier to yield a highly scalable and responsive discovery service.

\subsection{Effectiveness of Hierarchical Structure against Attention Dilution}
\label{subsec:hierarchical_acc}
\begin{figure}
    \centering
    \includegraphics[width=0.45\textwidth]{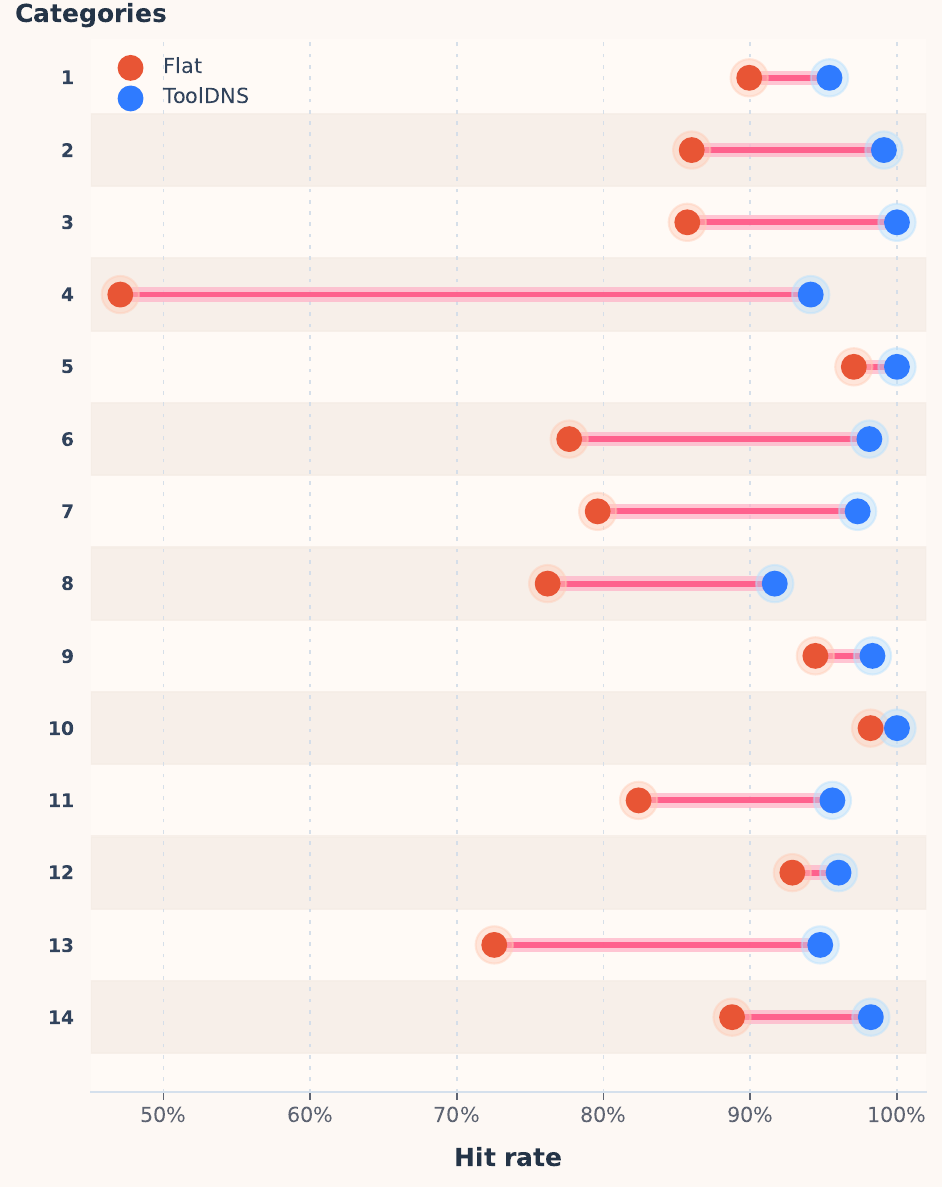}
    \caption{Retrieval accuracy of hierarchical versus flat organization across tool categories. The hierarchical approach consistently outperforms the flat baseline, demonstrating the value of structural pruning in mitigating attention dilution. A full enumeration of categories is provided in Appendix~\ref{app:tool_categories}.}
    \Description{Retrieval accuracy of hierarchical versus flat organization across tool categories. The hierarchical approach consistently outperforms the flat baseline, demonstrating the value of structural pruning in mitigating attention dilution. A full enumeration of categories is provided in Appendix~\ref{app:tool_categories}.}
    \label{fig:flatAcc}
\end{figure}

Having established the macro-level efficiency and accuracy of ToolDNS, we now turn to a more granular analysis of why the hierarchical design is so effective. The final experiment isolates a subtle yet critical advantage of the ToolDNS architecture: its ability to mitigate attention dilution in the LLM-based semantic matching modules deployed at authoritative servers. When an LLM is presented with a vast, flat list of candidate subdomains or tools, its attention can be scattered by irrelevant but superficially similar entries, degrading decision precision. We hypothesize that the stepwise delegation in ToolDNS acts as a semantic pruning mechanism, allowing the model to make a sequence of localized, high-confidence decisions rather than a single global comparison.

To test this hypothesis, we compared two organizational paradigms for presenting candidate information to the LLM scorer:
\begin{enumerate}[leftmargin=0.5cm]
\item \textit{Hierarchical resolution}: The model makes a step-wise decision, mimicking the ToolDNS delegation logic. At each level, it evaluates only the children of the current node.
\item \textit{Flat retrieval}: All leaf subdomains are expanded into a single, undifferentiated list. The model is asked to select the most relevant item from this exhaustive set.
\end{enumerate}

Crucially, we controlled for information content. In the flat scheme, each candidate was presented with its complete two-level hierarchical path information (e.g., \path{translate.nlp.tools}) to ensure that the model possessed the same contextual knowledge as it would acquire sequentially in the hierarchical scheme. The sole independent variable was the structure of presentation: bulk load versus progressive filtering. The System prompt used in this experiment is provided as `` Agent Role of NS server'' and ``Flat topK prompt'' in Appendix.\ref{app:prompts}

The results, shown in Figure~\ref{fig:flatAcc}, demonstrate that the hierarchical scheme consistently achieves a higher hit rate across all domains. This provides evidence that a hierarchical namespace is not merely an organizational convenience but a functional mechanism for improving LLM decision quality in hyper-scale search spaces. In the flat scheme, all candidates compete for attention simultaneously, introducing substantial semantic noise that can mislead the model. ToolDNS's hierarchical structure inherently implements a divide-and-conquer strategy. By evaluating only a small, semantically focused subset of nodes at each resolution step, the model effectively ignores the vast majority of irrelevant branches. This transformation from global attention competition to localized precision matching fundamentally alleviates attention dilution, enabling more accurate and reliable tool discovery at scale.

\section{Conclusion}
This work presents ToolDNS, a framework that explores an alternative path for AI tool discovery by repurposing the mature, globally deployed Domain Name System. The central insight emerging from this study is that for certain classes of discovery problems, particularly those emphasizing decentralized governance, low deployment barriers, and massive scale, infrastructure-native approaches merit renewed consideration alongside newer, above-application-layer indexing systems.

The impact of ToolDNS lies not in supplanting existing vector retrieval or context-injection methods, which remain effective for many use cases and deployment environments, but in expanding the design space. The framework demonstrates that by embracing the hierarchical semantics and delegation chains already present in DNS, one can achieve substantial reductions in search complexity and protocol overhead without requiring new global infrastructure. The logical subdomain model further illustrates how verifiable trust can be rooted in existing organizational identities rather than a single central authority, a property that becomes increasingly valuable as inter-agent collaboration crosses administrative boundaries.

Looking forward, we view ToolDNS as a complementary primitive within a broader discovery ecosystem. It may serve as a scalable entry point that routes queries to more specialized, high-precision registries, or as a fallback mechanism for agents operating under stringent trust or deployment constraints. The principles explored in this paper offer a set of reusable patterns that may inform the design of future systems. Ultimately, as the population of AI agents continues to grow, the most robust discovery fabric will likely emerge from a diversity of approaches, each addressing different points in the trade space between precision, cost, trust, and deployability. We hope that ToolDNS contributes one useful, interoperable thread to that fabric.

\appendix
\section{Prompt Templates}
\label{app:prompts}

This appendix presents the core prompt templates used in ToolDNS. Prompt A is used for semantic matching at TLD servers, while Prompt B is deployed on authoritative NS servers. In our experimental evaluation, Prompt A contributes to the accuracy calculation. Prompt B serves a dual purpose: it is used both for assessing retrieval accuracy and as the basis for comparison against the flat retrieval scheme, which utilizes Prompt C.

\begin{promptbox}[title=System Prompt A: Agent Role of TLD server]
You are a service-discovery ranking expert.

Task: Given a user **query**, a service domain list, select the top **\{k\}** domains from the candidate service domains list that can be used to solve user's query.

User query:
\{query\_description\}

Candidate service domains (each domain descripes a service's category and capability):
\{service\_list\}

Ranking rules:

1. Identify the core functionality in the of the Candidate service domains.

2. Select no more than \{k\} domains that are most critical for performing the query task.

3. If no domain is relevant, avoid forcing classifications for unclear entries and return \"other\".

4. The domain must be one of the candidate service domains.

5. Return domains in descending relevance order and splited by comma. No explanation, no markdown, no extra keys.
\end{promptbox}

\begin{promptbox}[title=System Prompt B: Agent Role of NS server]
You are a service-discovery ranking expert in domain \{topdomain\}.

Task: Given a user **task** and a service category list, select the top **\{k\}** category from the candidate service list that can be used to solve user's task.

User task:
\{task\_description\}

Candidate service category list (each category describes a service's category and capability):
\{service\_list\}

Ranking rules:
1. Identify the core functionality in the of the Candidate service category.
2. Select no more than \{k\} categories that are most critical for performing the task.
3. If no category is relevant, avoid forcing classifications for unclear entries and return \"other\".
4. The category must be one of the candidate service categories.
5. Return categories in descending relevance order and splited by comma. No explanation, no markdown, no extra keys.
\end{promptbox}

\begin{promptbox}[title=System Prompt C: Flat topK prompt]
\label{prompt:flat}
You are a service-discovery ranking expert in domain tools.
Task: Given a user **task** and a service category list, select the top **\{k\}** category from the candidate service list that can be used to solve user's task.

User task:
\{task\_description\}

Candidate service category list (each category describes a service's category and capability):
\{service\_list\}

Ranking rules:
1. Identify the core functionality in the of the Candidate service category.
2. Select no more than \{k\} categories that are most critical for performing the task.
3. If no category is relevant, avoid forcing classifications for unclear entries and return \"other\".
4. The category must be one of the candidate service categories.
5. Return categories in descending relevance order and splited by comma. No explanation, no markdown, no extra keys.
\end{promptbox}

\section{List of Tool Categories}\label{app:tool_categories}
This appendix enumerates the hierarchical tool categories defined in our dataset. These categories serve as the evaluation taxonomy for the experiments described in Sections \ref{subsec:hit_rate} and \ref{subsec:hierarchical_acc}.

\begin{enumerate}
  \item Web\_Search\_and\_SEO.
  \item IoT.
  \item Religion\_and\_Spirituality.
  \item Agriculture\_or\_Horticulture.
  \item Logistics.
  \item Blockchain.
  \item Coding.
  \item Calendar.
  \item Social\_Media.
  \item Weather\_and\_Climate.
  \item Video.
  \item Security.
  \item Email.
  \item Audio.
\end{enumerate}

\bibliographystyle{ACM-Reference-Format}
\bibliography{references}

\end{document}